\documentclass[runningheads]{llncs}
\usepackage{graphicx}
\usepackage{amsmath}
\usepackage{amssymb}
\usepackage{mathtools}
\usepackage{amsthm}
\usepackage{bm}
\usepackage{setspace}
\usepackage{multirow}
\usepackage{hyperref}

\usepackage{amsmath}
\usepackage{algorithm}
\RequirePackage{algorithmic}
\usepackage{hyperref}
\usepackage{xcolor}

\usepackage{caption}
\usepackage{subcaption}

\begin{document}
\title{Identifying the Source of Generation for Large Language Models}
%
%
\author{Bumjin Park \inst{1}\orcidID{0000-0002-4568-2813} \and
Jaesik Choi\inst{1,2}\orcidID{0000-0002-4663-3263} }
\authorrunning{Park, B. and Choi, J.}
%
\institute{Graduate School of AI, KAIST \and
Corresponding Author \\ 
\email{\{bumjin,jaesik.choi\}@kaist.ac.kr}}
\maketitle              
\begin{abstract}
Large language models (LLMs) memorize text from several sources of documents. In pretraining, LLM trains to maximize the likelihood of text but neither receives the source of the text nor memorizes the source. Accordingly, LLM can not provide document information on the generated content, and users do not obtain any hint of reliability, which is crucial for factuality or privacy infringement.  This work introduces token-level source identification in the decoding step, which maps the token representation to the reference document. We propose a bi-gram source identifier, a multi-layer perceptron with two successive token representations as input for better generalization. We conduct extensive experiments on Wikipedia and PG19 datasets with several LLMs, layer locations, and identifier sizes. The overall results show a possibility of token-level source identifiers for tracing the document, a crucial problem for the safe use of LLMs.  Code is available at \href{https://github.com/fxnnxc/source_identification_of_llm}{https://github.com/fxnnxc/DocSourceLLM}. 


\keywords{Large Language Models \and Document Source Identification  \and Probing \and Copyright Protection }
\end{abstract}

\definecolor{color0}{rgb}{1.0, 1, 0.5}
\definecolor{color1}{rgb}{1.0,0.5, 0.6}
\definecolor{color2}{rgb}{0,1,0.5}
\definecolor{color3}{rgb}{0.0,0.5, 1.0}
\definecolor{no}{rgb}{1.0, 1.0, 1.0}

\section{Introduction}

Recent advances in large language models (LLMs) show human-level performance, and LLM has been used in several applications such as chatbot systems, management systems, medical AIs\cite{johnson2023assessing,taecharungroj2023can}. Concurrently, the impact of LLMs on society has increased, and several concerns and regulations on LLMs have increased.
Researchers discussed safety issues of LLMs worries such as false information, social bias, privacy infringement\cite{10198233}, and hallucinations\cite{carlsmith2023scheming,10.1145/3593013.3594067,shoker2023confidencebuilding}. One need for LLMs is to provide the reasons for generated content. LLMs can provide rich information by generating longer sentences and descriptions, but the information still depends on the generated content. Another way of explanation is to provide the source of generated content, the original documents. 

However, providing the source of information encounters several limitations. One problem is that the generated content mainly consists of multiple words from different sources; even a single word could originate from multiple documents. For example, the source of the sentence \textit{"apple is delicious."} would be multiple documents. Second, the generated content can consist of multiple documents. For example, \textit{"Apple is a company and is delicious."} is a mixture of contents in two documents. These observations reveal that identifying the source is a multi-label prediction and word-level identification problem. 

Unlike LLMs, when humans generate content related to factual knowledge, they can provide the learned material and use additional information in the document. For example, the source of the sentence \textit{"queen offers Snow White a poisoned [MASK]."} can be inferred from the book \textit{"Snow White" and the "[MASK]"} location could be inferred as \textit{"apple"}. Based on this idea, the work assumes that the internal representation of LLMs and the prediction of the next word could reveal the origin of the sentences. To verify this, we investigate the possibility of source tagging at the token level as a multi-label prediction and word-level identification.

\begin{figure}[h]
\centering
\includegraphics[width=12cm]{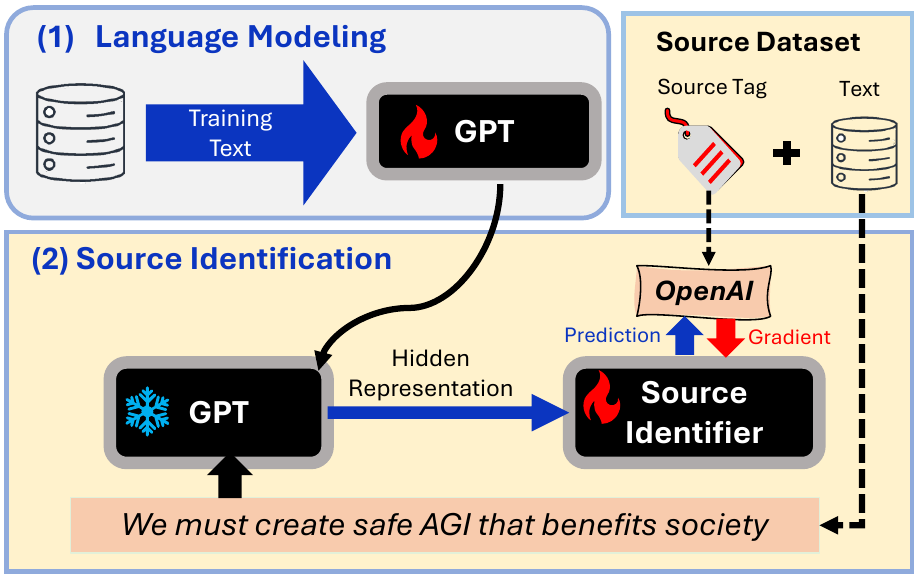}
\caption{An illustration of the source identification. In step (1), the GPT model is trained with language modeling to memorize documents. In step (2), the source identifier is trained to predict the documents while the GPT is frozen. Note that the pretraining data is a collection of multiple documents with known sources. }
\label{source_identification_problem}
\end{figure}

In detail, we focus on the fact that the pretraining stage includes the data sources, and at least we can tag the documents in the pretraining to match the source. Figure \ref{source_identification_problem} illustrates the source tagging problem. In pretraining, GPT memorizes texts. In the phase of source identification, a source identifier maps text to the source. The parameters of GPT are fixed, and the original forward process of the model is not modified. The source identification process maps the internal representation of GPT and provides the label of the reference documents. Table \ref{table:example1} shows an example of the source identification.

\begin{table}[]
    \centering
    \small
    \def\arraystretch{1.0}
    \caption{A passage generated by Llama2 with the prompt "Christmas" and the source identification for 100 documents in Wikitext-103-v1. Numbers in blanks are the predicted labels. Chrismas-related words mainly originated from Wiki pages, \textit{Chrismas 1994} and \textit{New Year's Eve.}}
\begin{tabular}{|l|} 
\noalign{\hrule height 1pt}
Christmas is a time for family, friends, and food. \\ 
It’s a time to celebrate the birth of Jesus Christ and the joy of the season. \\
\noalign{\hrule height 1pt}
\colorbox{color0!30}{\textunderscore Christmas} \colorbox{color0!30}{\textunderscore is} \colorbox{color0!30}{\textunderscore a} \colorbox{color1!30}{\textunderscore time} \colorbox{color1!30}{\textunderscore for} \colorbox{color0!30}{\textunderscore family} \colorbox{color0!30}{,} \colorbox{no!30}{\textunderscore friends} \colorbox{color0!30}{,} \colorbox{color3!30}{\textunderscore and} \colorbox{color2!30}{\textunderscore food} \\
\colorbox{color0!30}{.} \colorbox{color0!30}{\textunderscore It} \colorbox{no!30}{’} \colorbox{no!30}{s} \colorbox{color0!30}{\textunderscore a} \colorbox{color1!30}{\textunderscore time} \colorbox{color1!30}{\textunderscore to} \colorbox{no!30}{\textunderscore celebr} \colorbox{no!30}{ate} \colorbox{color2!30}{\textunderscore the} \colorbox{no!30}{\textunderscore birth} \colorbox{no!30}{\textunderscore of} \colorbox{no!30}{\textunderscore Jesus} \\
\colorbox{color1!30}{\textunderscore Christ} \colorbox{no!30}{\textunderscore and} \colorbox{no!30}{\textunderscore the} \colorbox{no!30}{\textunderscore joy} \colorbox{no!30}{\textunderscore of} \colorbox{color2!30}{\textunderscore the} \colorbox{color2!30}{\textunderscore season} \colorbox{color0!30}{.}
\\
\hline 
\colorbox{color0!30}{Document(47):Christmas 1994 nor 'easter } \\
\colorbox{color1!30}{Document(10):Wins} \\ 
\colorbox{color2!30}{Document(99):New Year 's Eve ( Up All Night ) } \\
\colorbox{color3!30}{Document(46):Wrapped in Red } \\
  \noalign{\hrule height 1pt}
    \end{tabular}
    \label{table:example1}
\end{table}

This paper introduces a token-level source identification problem and proposes to utilize a multi-layer perceptron to map token representation and the source of the documents. We study the trainability of relationships on source, sentences, and LLMs and show that scalability holds for source tagging. This paper aims to ensure the safe usage of LLMs by tagging the source of the generated contents. We summarize our contribution to the token-level source identification as follows.  
\begin{itemize}
    \item We formalize the token-level source identification as a multi-label prediction. 
    \item We propose a bi-gram prober for better generalization. 
    \item We investigate several LLM models (Pythia \cite{biderman2023pythia}, OPT\cite{zhang2022opt}, and Llama2 \cite{touvron2023llama}), sizes (70M to 13B), layers (the first layer to logits).
\end{itemize}

\section{Related Work}

\subsection{Text Provenance} 
Tracing text provenance is crucial to protect copyright or intellectual property (IP) \cite{yang2021tracing}. Several works contribute to the text provenance of LLMs. Membership inference determines the misuse of user information in LLM 
 \cite{shokri2017membership,hisamoto2020membership}. Backdoor attacks hide triggers to verify the use of private information. Watermark methods are applied to trace text provenance or guarantee the machine-generated contents \cite{zhao2023provable,pmlr-v202-kirchenbauer23a}. Most of these works assume an adversarial relationship between LLMs and users. On the other hand, this work assumes a cooperative relationship between LLM and owners of documents as the source identification training can provide the referenced documents. 

\subsection{Belifs on LLMs and Probers}
The source identifier finds referenced documents from representations in LLMs rather than a generated word.
Getting the document label from a token representation is based on the belief that LLM can acknowledge the source information when generating content. LLMs, also black block models, have many neurons that possibly represent syntactic or semantic meanings \cite{cammarata2020thread}. Probing the neurons is done by training a classifier to map the representations and labels \cite{kim2018interpretability}. Recent work on LLMs shows that LLMs know the factuality of the contents \cite{azaria2023internal,agrawal2023language}. Although this belief is disputable because of spurious correlation \cite{levinstein2023still}, one consensus is that probers provide separable labels from representations when they are trained \cite{kim2018interpretability}. The source identification problem leverages the beliefs of LLMs and the ability of probers.

\section{Methods}

\subsection{Token-level Source Identification}

\begin{figure}[h]
\centering
\includegraphics[width=11cm]{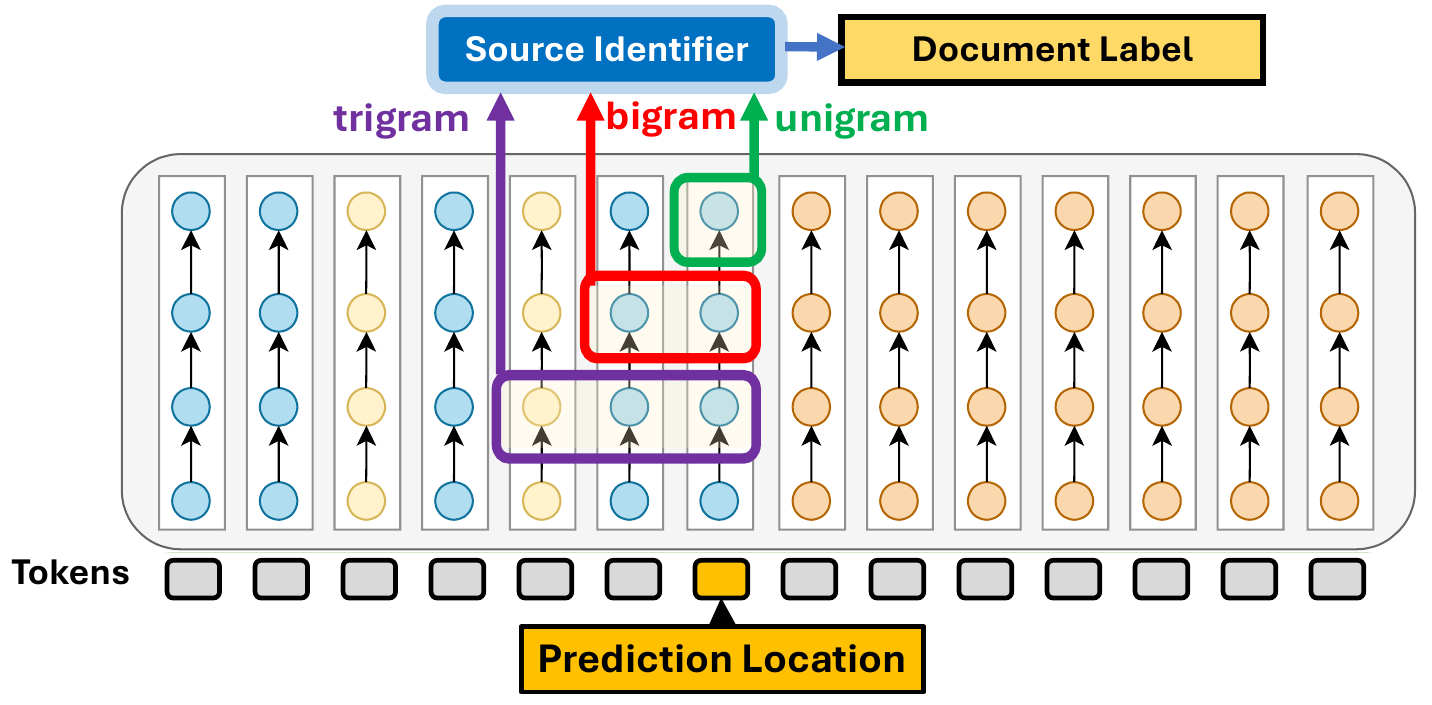}
\caption{Dataset construction and $n$-gram source identifier. The current prediction location is \textit{train}, and the source identifier uses $n$-gram representations as inputs.}
\label{n_gram_prediction}
\end{figure}

As we discussed in the introduction, the prediction of documents is a multilabel prediction problem. In addition, when the number of documents is huge, such as 1M, the problem is extreme multilabel classification (XML) \cite{dahiya2021deepxml}. Following the previous XML works, we use the binary cross entropy loss to train MLP to predict document labels.  For the dataset construction, our token-level prediction has the following 3 splits: \textit{train}, \textit{test-in}, and \textit{test-out}. We first select train tokens randomly and other tokens in between train tokens are \textit{test-in}. The \textit{test-out} has tokens outside of the train tokens. These splits are required to test the generalization of source identification rather than memorization, as non-linear MLP can train even noisy inputs. 
Figure \ref{n_gram_prediction} shows the locations of three splits.

\subsection{$n$-gram Source Identifier}

In this section, we describe the $n$-gram source identifier $f$, which predicts document labels from the representation of document texts. 
The source identifier is multi-layers MLP with ReLU activation following previous MLP prober designs for LLMs \cite{li2022emergent}. 
We use notations $y = (y_1,y_2, \cdots, y_t)$ for texts of length $t$, in document $D$ and $h^{\ell} = (h_{1}^{\ell}, h_{2}^{\ell}, \cdots, h_{t}^{\ell}) $  for the internal representations of $y$ at layer $\ell$. The $n$-gram source identifier $f$ predicts the document $D$ at token location $t'$ from $n$-gram representations $h^{\ell}_{t'-n:t'}$
\begin{equation}
\hat{D} = f(h^{\ell}_{t'-n:t'})
\end{equation}
where $\hat{D}$ is the predicted document from hidden representations $h^{\ell}_{t'-n:t'}$ with $f$. 
Figure \ref{n_gram_prediction} shows the $n$-gram representations and the source identifier. 
This work uses $n=\{1,2,3\}$ to verify the separability of documents from $n$-gram representations. Although we can increase the $n$ by more than three, most three-gram representations are already sparse, and bigram is enough complexity\footnote{When the number of tokens is 50K, the complexity of bigram is 50K $\times $50K. We assume similar complexity for the internal representations.}. The choice of layer location $\ell$ is an inductive bias for the document separation. A higher layer would be better when lexical words can separate the documents. 

 \subsection{Training of $n$-gram Source Identifier}

Training source identifier has two stages: (1) gather activation of LLMs for documents and (2) minimize entropy loss for document indices. As the source identification is a multi-label, binary cross-entropy loss (BCE) is better than the cross-entropy loss (CE) to stabilize the training\footnote{CE normalizes the probability for all documents, while BCE computes logits for individual document entries.}. Algorithm \ref{alg:training} shows the training procedure. When we map the document from logits, we gather the logits of document texts.  

\begin{algorithm}[h]
   \caption{Training Source Identifier with Internal Activations}
   \label{alg:training}
\begin{algorithmic}[1]
  \STATE {\bfseries input:} pretrained LLM, source identifier $f$, target layer $\ell$, document indices $d=1,2,\cdots,N$, text length $t$, and training epochs $K$.
\STATE gather internal activation $H_{d} = \{h^{\ell}_1, h^{\ell}_2, \cdots, h^{\ell}_t \}$ with LLM for all document indices $d=1,2,\cdots,N$.
\FOR{epochs in $1,2,\cdots, K$}
\FOR{document $d$ in $1,2,\cdots, N$}
\STATE update $f$ with $(H_{d}, d)$ to minimize BCE loss. 
\ENDFOR
\ENDFOR
\end{algorithmic}
\end{algorithm}

 \clearpage 

\section{Experiments}

We train source identifiers for  Llama2\cite{touvron2023llama} (7b,13b, 7b-chat, 13b-chat), Pythia\cite{biderman2023pythia} (410m, 1.4B, 6.9B, 12B), OPT\cite{zhang2022opt}(350M, 2.7B, 6.7B, 13B)  for first 100 documents in Wikitext-103-v1 and PG19 datasets. The MLP size of source identifiers is either linear or non-linear models (\textit{tiny, small, medium, and large}) with ReLU activation. The hidden size of \textit{tiny} is 128, and \textit{large} has (128, 256, 512, 1024) hidden sizes. We use AdamW optimizer with 0.001 learning rate with 64 batch size. We gathered hidden 512 token representations of LLMs and split them into 180, 76, and 256 for train, test-in, and test-out, respectively. 

\subsection{Identification Accuracy and Size of Large Language Models}

\begin{figure}[h]
\centering
\includegraphics[width=\textwidth]{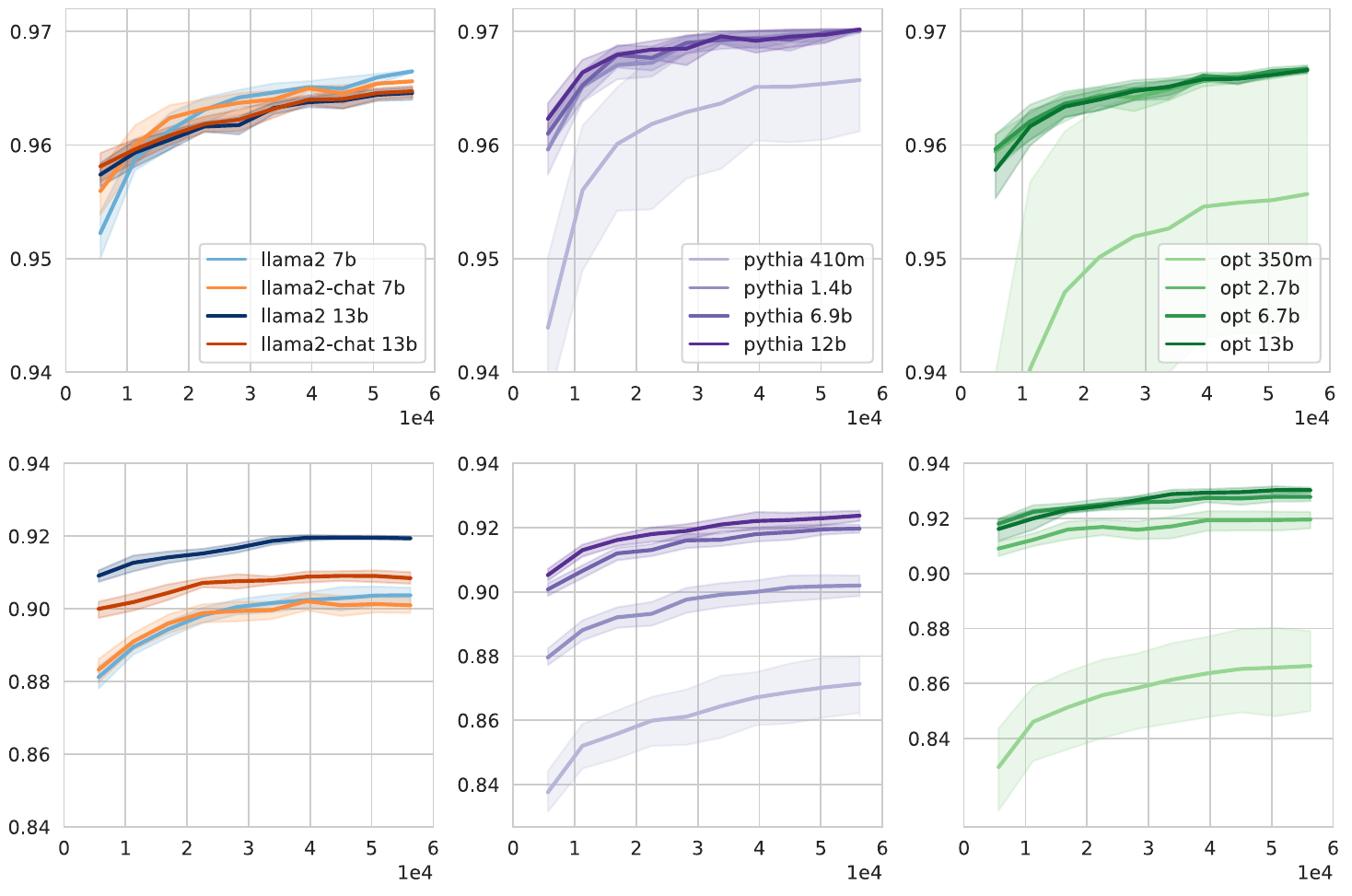}
\caption{Training accuracy of three model types and sizes. The first row is the accuracy for \textit{train}, and the second is for \textit{test-in}. Larger models show better generalization performance. In the case of Llama2, 7B size shows higher \textit{train} accuracy than 13B but shows better generalization with 13B. }
\label{llm_scalability}
\end{figure}

We first evaluate the performance of unigram and medium-size source identifiers depending on the size of LLMs. The prediction accuracy for \textit{train} and \textit{test-in} is shown in Figure \ref{llm_scalability}. As the size of LLM increases, the prediction accuracy for \textit{train} and \textit{test-in} both increases. However, although LLMs show similar performance for large sizes, \textbf{the generalization performance is better for large language models}. 

\subsection{Layers of Large Language Models}

\begin{figure}[h!]
    \captionsetup[subfigure]{justification=centering}
     \centering
     \begin{subfigure}[b]{\textwidth}
         \centering
         \includegraphics[width=9.2cm]{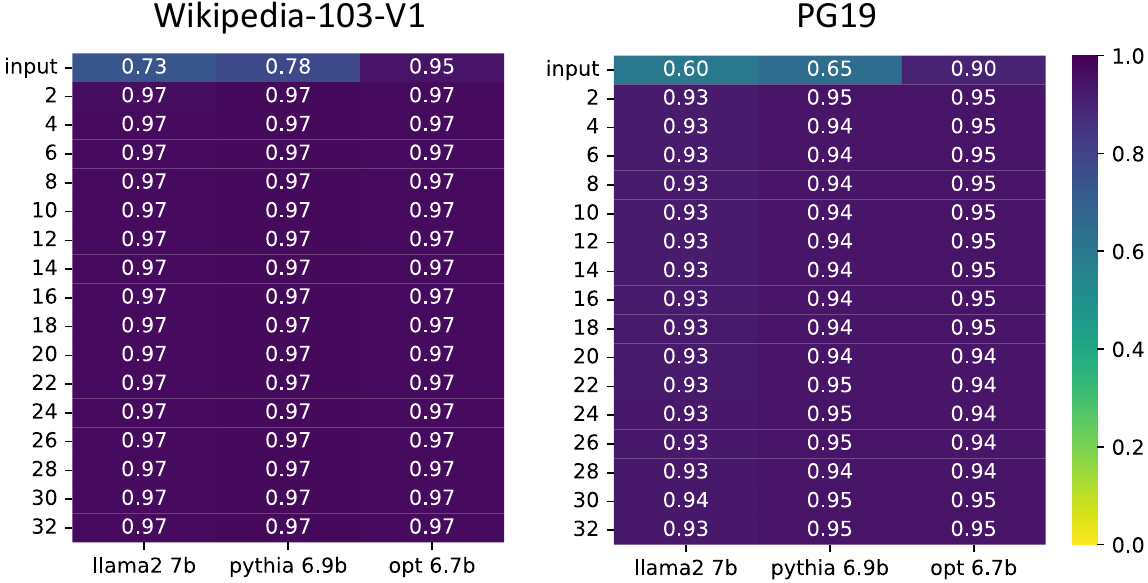}
         \caption{\textit{train} accuracy}
         \label{location_train}
     \end{subfigure}
     \hfill
     \begin{subfigure}[b]{\textwidth}
         \centering
         \includegraphics[width=9.2cm]{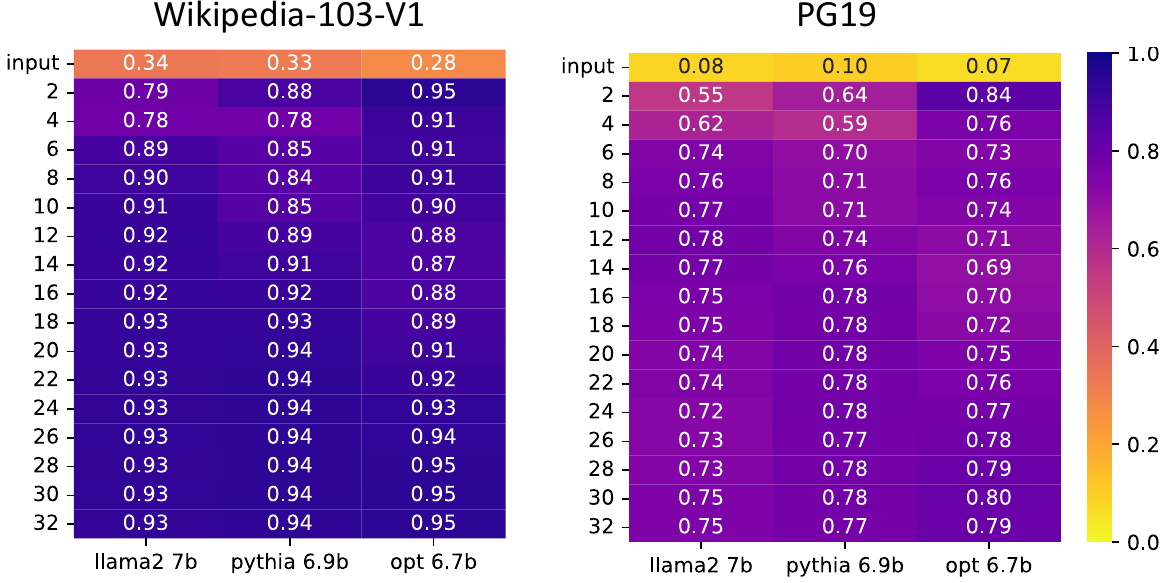}
         \caption{\textit{test-in} accuracy}
         \label{location_eval}
     \end{subfigure}
     \caption{Accuracy of \textit{train} and \textit{test-in} splits. Although \textit{train} dataset shows almost the same accuracy over layers, the generalization differs in layers for \textit{test-in}.}
\end{figure}

Previous works show that the lower layer captures syntactic information, and the upper layer has semantic information \cite{geva2021transformer}. 
A known assumption is that the last representation is related to the logits, the next word prediction, and the internal representation captures the semantic meaning of sentences. We train bigram medium-size source identifiers for Wikitext and PG19 with internal layers to verify this assumption. Figure \ref{location_train} and \ref{location_eval} show the accuracy for train and test-in, respectively. Almost all layers show similar training performance ($97\%$), while the test-in differs over layers. We observe that \textbf{the best performance of Wikitext is obtained in the last layer}, while \textbf{PG19 is in the middle layers}. These results support our assumption because the documents in Wikitext could be separable from words, while documents in PG19 are book contents and could be separable with semantic representation. Therefore, we must consider the separable features of documents, such as words, sentences, or semantics, and choose the layer considering the separable features in the documents. 

\subsection{$n$-Gram Comparison}

\begin{figure}[h]
\centering
\includegraphics[width=12cm,trim={0cm 0.2cm 0.0cm 0},clip]{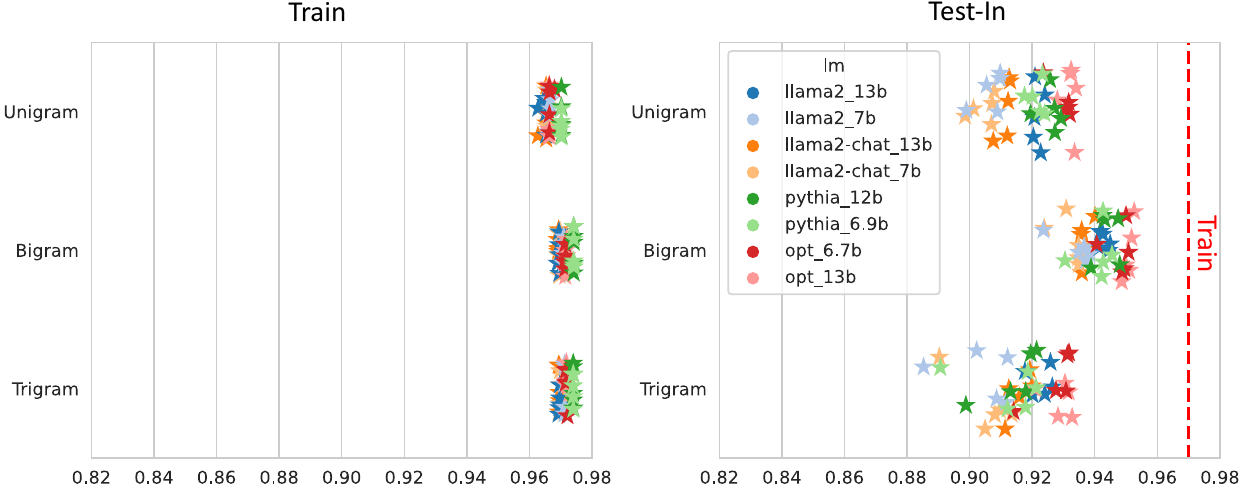}
\caption{Accuracy of \textit{train} and \textit{test-in} splits. Each star represents a single MLP. All $n$-gram shows the same \textit{train} performance, but the bigram generalizes better. }
\label{gram_comparison}
\end{figure}

In this section, we compare $n$-gram representations. 
 Figure \ref{gram_comparison} shows the final prediction accuracy for training and test-in for all MLP sizes with Wikitext-103-v1, respectively. We observe that the $n$-gram has no effect in training but has a significant gap for generalization. The bigram representation shows better performance than unigram and trigram representations. This observation is because the unigram is insufficient, while the trigram is too complex to separate documents. Note that as $n$ increases, the complexity of the space increases exponentially. The results indicate that \textbf{bigram representation complexity is the choice for the separability of documents.}

\begin{figure}[h]
\centering
\includegraphics[width=12cm, trim={0cm 0.2cm 0.0cm 0},clip]{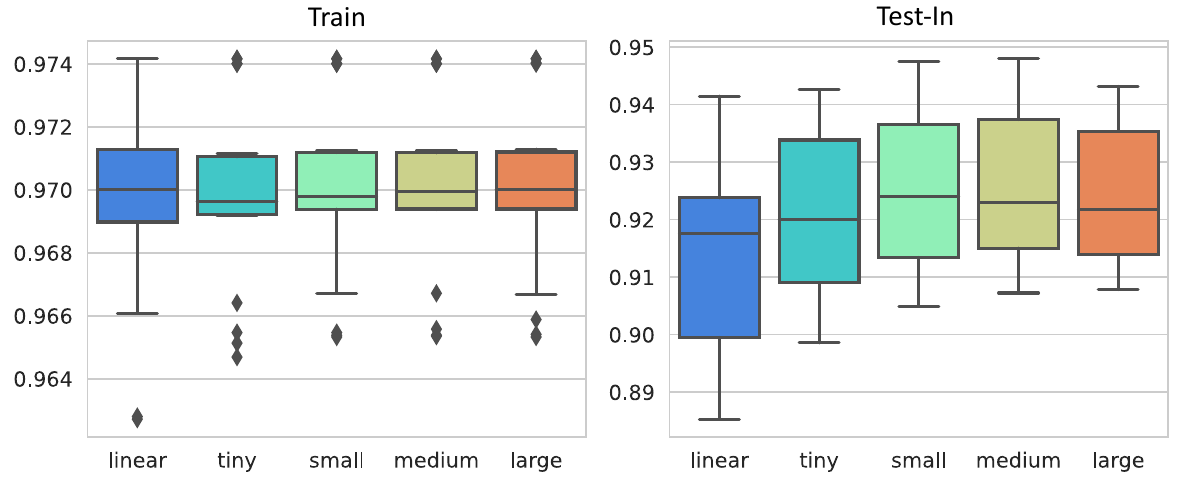}
\caption{MLP size comparison. Each point is different $n$-gram and seed result. Small and medium model sizes generalize better for source identification.  }
\label{mlp_comparison}
\end{figure}

\subsection{Size of Source Identifiers}
The size of an MLP also affects the source identification performance. In the case of a linear model, it separates the documents based on the activated neuron in LLM. In contrast, non-linear MLP assumes a non-linear relation of neuron activation on the separation of documents. Figure \ref{mlp_comparison} shows the train and test-in performances for bigram source identifiers with all model sizes. The accuracy distribution is the same for all model sizes, but the generalization differs. The small and medium sizes show the best generalization. The linear model shows the lowest performance. Note that a single MLP predicts the mapping of all tokens in documents. Therefore, non-linear, known to cut the manifold, shows better generalization. 

The model size is related to the document length and the number of documents. As the pre-training data size increases, the size of MLP must increase to cover all token representations.

\subsection{Qualtitative Identification Analysis}

The source identifier is a multi-label predictor. To visualize the prediction results in a human-understandable manner, we highlight tokens whose probability is higher than 0.99 and take the label with the maximum document. We identify the source of tokens for 100 documents of Wikitext-103-v1 and PG19. Tables \ref{table:example1}, \ref{table:example2}, and \ref{table:example3} show the quantitative examples of generated contents from Llama2-7B and medium-size bigram source identifiers.

\begin{table}[h]
    \centering
    \small
    \def\arraystretch{1.0}
\caption{Generation example of Llama2-7B with a prompt \textit{"Shipwrecked"} and the source identification for 100 documents in PG19. The first row is the generated passage, and the second is the identified documents with a threshold cut of 0.99. The word \textit{"Robinson"} exists in the book \textit{"Mystery"}. However, \textit{Crusoe} is not. Hence, the identification is limited to the candidate documents.}
\begin{tabular}{|l|} 
\noalign{\hrule height 1pt}
Shipwrecked on a desert island, Robinson \\ Crusoe is forced to make do with only a knife, a gun and an empty bottle. \\ 
But he is not alone on his island paradise. He has company  \\
\hline 
'\colorbox{no!30}{} \colorbox{no!30}{} \colorbox{no!30}{ \textunderscore Sh} \colorbox{no!30}{ip} \colorbox{color3!30}{w} \colorbox{no!30}{reck} \colorbox{no!30}{ed} \colorbox{no!30}{\textunderscore on} \colorbox{no!30}{\textunderscore a} \colorbox{color0!30}{\textunderscore desert} \colorbox{color0!30}{\textunderscore island} \colorbox{no!30}{,} \colorbox{color3!30}{\textunderscore Robinson} \\ \colorbox{no!30}{\textunderscore Cr} \colorbox{no!30}{us} \colorbox{no!30}{oe} \colorbox{no!30}{\textunderscore is} \colorbox{no!30}{\textunderscore forced} \colorbox{no!30}{\textunderscore to}  \colorbox{no!30}{\textunderscore make} \colorbox{no!30}{\textunderscore do} \colorbox{no!30}{\textunderscore with} \colorbox{no!30}{\textunderscore only} \colorbox{no!30}{\textunderscore a} \colorbox{color1!30}{\textunderscore kn} \colorbox{color1!30}{ife} \\ \colorbox{no!30}{,} \colorbox{no!30}{\textunderscore a} \colorbox{color1!30}{\textunderscore gun} \colorbox{no!30}{\textunderscore and} \colorbox{no!30}{\textunderscore an} \colorbox{no!30}{\textunderscore empty} \colorbox{no!30}{\textunderscore bott} \colorbox{color0!30}{le} \colorbox{no!30}{.} \colorbox{no!30}{} \colorbox{no!30}{But} \colorbox{no!30}{\textunderscore he} \\ \colorbox{no!30}{\textunderscore is} \colorbox{no!30}{\textunderscore not} \colorbox{color2!30}{\textunderscore alone} \colorbox{color2!30}{\textunderscore on} \colorbox{no!30}{\textunderscore his} \colorbox{no!30}{\textunderscore island} \colorbox{no!30}{\textunderscore parad} \colorbox{no!30}{ise} \colorbox{no!30}{.} \colorbox{no!30}{\textunderscore He} \colorbox{no!30}{\textunderscore has} \colorbox{color2!30}{\textunderscore company }' \\ 
\hline 
\colorbox{color0!30}{Document(9):Wild Northern Scenes. } \\ 
\colorbox{color1!30}{Document(30):The Hunt Ball Mystery} \\ 
\colorbox{color2!30}{Document(5): A Voyage to the Moon by George Tucker} \\ 
\colorbox{color3!30}{Document(8): The Mystery by Stewart Edward White and Samuel Hopkins Adams} \\ 
  \noalign{\hrule height 1pt}
    \end{tabular}
    \label{table:example1}
\end{table}

\begin{table}[h]
    \centering
    \small
    \def\arraystretch{1.0}
\caption{Generation example of Llama2-7B with a prompt \textit{"The Book"}, and the source identification for 100 documents in Wikitext. The generation is related to the funerary and magic spells. Most semantic and lexical tokens are identical to words in the \textit{"Ancient Egyptian deities"} wiki page.}
\begin{tabular}{|l|} 
\noalign{\hrule height 1pt}
The Book of the Dead is a collection of funerary texts from a variety of sources \\ dating from the 16th to 11th century BC. \\ The collection includes spells, magic formulas, and prayers of funeral \\
\hline 
\colorbox{no!30}{ \textunderscore The} \colorbox{no!30}{\textunderscore Book} \colorbox{no!30}{\textunderscore of} \colorbox{no!30}{\textunderscore the} \colorbox{no!30}{\textunderscore Dead} \colorbox{no!30}{\textunderscore is} \colorbox{no!30}{\textunderscore a} \colorbox{color0!30}{\textunderscore collection} \colorbox{color0!30}{\textunderscore of} \colorbox{color0!30}{\textunderscore fun} \colorbox{color0!30}{er} 
\colorbox{color0!30}{ary} \colorbox{color0!30}{\textunderscore texts}  \\
\colorbox{color0!30}{\textunderscore from} \colorbox{color0!30}{\textunderscore a} \colorbox{color0!30}{\textunderscore variety} \colorbox{color0!30}{\textunderscore of} \colorbox{color0!30}{\textunderscore sources} \colorbox{no!30}{\textunderscore d} \colorbox{no!30}{ating} \colorbox{color0!30}{\textunderscore from} \colorbox{color0!30}{\textunderscore the} \\
\colorbox{color0!30} {\textunderscore } \colorbox{color0!30}{1} \colorbox{no!30}{6} \colorbox{no!30}{th} \colorbox{color0!30}{\textunderscore to} \colorbox{color0!30}{\textunderscore } \colorbox{color0!30}{1} \colorbox{no!30}{1} \colorbox{no!30}{th} \colorbox{no!30}{\textunderscore century} \colorbox{color0!30}{\textunderscore BC} \colorbox{color0!30}{.} \colorbox{color0!30}{\textunderscore The} \colorbox{color0!30}{\textunderscore collection} \colorbox{color0!30}{\textunderscore includes} \\
\colorbox{color0!30}{\textunderscore sp} \colorbox{color0!30}{ells} \colorbox{color0!30}{,} \colorbox{color0!30}{\textunderscore magic} \colorbox{color0!30}{\textunderscore formulas} \colorbox{color0!30}{,} \colorbox{color0!30}{\textunderscore and} \colorbox{color0!30}{\textunderscore pray} \colorbox{color0!30}{ers} \colorbox{color0!30}{\textunderscore of} \colorbox{color0!30}{\textunderscore fun }'
\\
\hline 
\colorbox{color0!30}{Document(20):Ancient Egyptian deities } \\ 
  \noalign{\hrule height 1pt}
    \end{tabular}
    \label{table:example2}
\end{table}


\begin{table}[h!]
    \centering
    \small
    \def\arraystretch{1.0}
\caption{Generation example of Llama2-7B with a prompt \textit{"ChatGPT"} and the source identification for 100 documents in PG19. The sentences describe several benefits of ChatGPT. The source identifier provides four documents, which may include a similar token representation. Note that the generation provides the most probable sentence, yet the source of contents may be various documents.}
\begin{tabular}{|l|} 
\noalign{\hrule height 1pt}
ChatGPT is a powerful tool that can help you create a better resume. \\ It can help you identify your skills and experience, \\ and it can also help you write a more effective resume. \\ 
\hline 
\colorbox{no!30}{} \colorbox{no!30}{} \colorbox{no!30}{ \textunderscore Ch} \colorbox{no!30}{at} \colorbox{no!30}{G} \colorbox{no!30}{PT} \colorbox{no!30}{\textunderscore is} \colorbox{no!30}{\textunderscore a} \colorbox{color0!30}{\textunderscore powerful} \colorbox{color0!30}{\textunderscore tool} \\ \colorbox{no!30}{\textunderscore that} \colorbox{no!30}{\textunderscore can} \colorbox{no!30}{\textunderscore help} \colorbox{no!30}{\textunderscore you} \colorbox{no!30}{\textunderscore create} \colorbox{no!30}{\textunderscore a} \colorbox{no!30}{\textunderscore better} \colorbox{color1!30}{\textunderscore res} \colorbox{no!30}{ume} \colorbox{no!30}{.}  \\ 
\colorbox{color0!30}{\textunderscore It} \colorbox{no!30}{\textunderscore can} \colorbox{no!30}{\textunderscore help} \colorbox{no!30}{\textunderscore you} \colorbox{color1!30}{\textunderscore identify} \colorbox{color1!30}{\textunderscore your} \colorbox{color0!30}{\textunderscore skills} \colorbox{color0!30}{\textunderscore and} \colorbox{color0!30}{\textunderscore experience} \colorbox{color1!30}{,} \colorbox{color0!30}{\textunderscore and} \\ \colorbox{no!30}{\textunderscore it} \colorbox{no!30}{\textunderscore can} \colorbox{no!30}{\textunderscore also} \colorbox{color2!30}{\textunderscore help} \colorbox{no!30}{\textunderscore you} \colorbox{no!30}{\textunderscore write} \colorbox{no!30}{\textunderscore a} \colorbox{color3!30}{\textunderscore more} \colorbox{no!30}{\textunderscore effective} \colorbox{no!30}{\textunderscore res} \colorbox{no!30}{ume} \colorbox{no!30}{. }
\\
\hline 
\colorbox{color0!30}{Document(10): Divine Comedy Inferno by Dante Alighieri} \\ 
\colorbox{color1!30}{Document(29): Spalding's Official Baseball Guide} \\ 
\colorbox{color2!30}{Document(72): Mobilizing Woman-Power } \\ 
\colorbox{color3!30}{Document(50): Divine Comedy Hell} \\ 
  \noalign{\hrule height 1pt}
    \end{tabular}
    \label{table:example3}
\end{table}

\subsection{Additional Studies}

In the previous experimental sections, we show that the source identifiers can separate documents with generalization. However, to trust the possibility of source identifiers, we must verify the source identifiers in (1) an extreme number of labels and (2) generalization for the \textit{test-out} tokens. We provide additional experimental results to verify the possibility of source identifiers. In addition, we compare the logits and the last hidden representations to show that the last hidden representation is a better choice than logits.

\subsubsection{Extreme Labels}
\begin{figure}[h!]
\centering
\includegraphics[width=10.0cm]{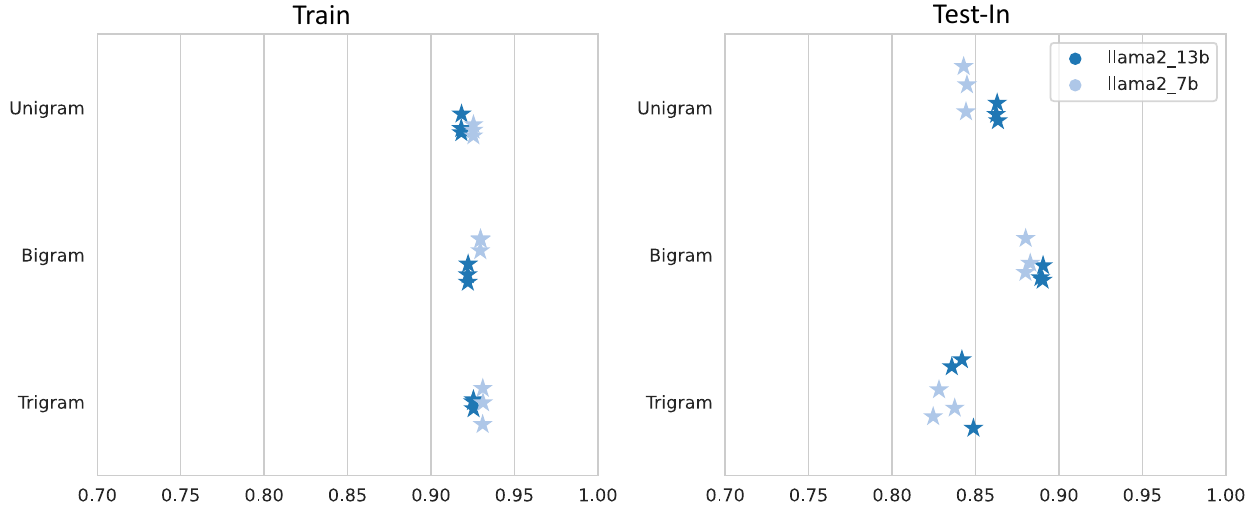}
\caption{Accuracy for all MLP sizes and 1K documents. The bigram case shows better generalization than other $n$-grams. In addition, the larger models generalize better. These results are consistent with the 100 document case (Figure \ref{gram_comparison}).}
\label{gram_comparison1000}
\end{figure}

In pretraining, the number of documents is huge. For example, Wikitext-103-v1 and PG19 both have more than 20K documents. To cover all the documents, we must verify the scalability of source identifiers for an extreme number of documents. Figure \ref{gram_comparison1000} shows the performance of medium-size prober with N-gram representations for 1K Wikitext documents. The generalization of bigram representation for 1K documents consists of the previous results. 

\begin{figure}[h!]
\centering
\includegraphics[width=10.0cm]{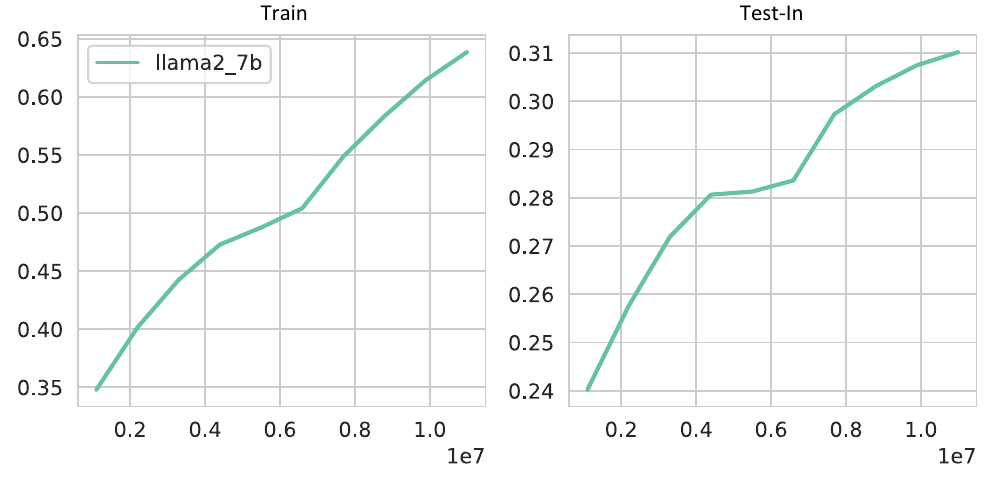}
\caption{Training dynamics of a large MLP for 10K documents and 1M training steps.  More resources are required to cover a huge number of documents. In addition, the generalization performance is almost half of the \textit{train}.   }
\label{extreme_label}
\end{figure}

We also check the training dynamics for 10K documents. 
Figure \ref{extreme_label} shows the training dynamics for 10K documents with large-size MLP. Note that train and test-in accuracies linearly increase as we train more, but the evaluation performance is almost half ($ 31\%$ compared to $62\%$). We believe that XML studies can train the source identifiers efficiently.

\subsubsection{Generalization on Test-out}

In actual use cases, the source identification must be trained for all the pre-training passages, and no out-of-distribution would be required. However, to verify the generalization, we evaluate the \textit{test-out} tokens, which may have semantic or syntactic similarities to learned \textit{train} tokens. Table \ref{table:test_out} shows \textit{test-out} accuracy for 100 Wikitext documents with medium-size source identifiers. The performance drop is $10 \sim 20\%$ for unigram and bigram cases, and trigram shows almost $30\%$ drop. 
The bigram model achieves the best \textit{test-out} performance. Thus, we conclude that bigram is the best $n$-gram representation.
In addition, the 13B model size shows $0.8$ accuracy compared to $0.77$ accuracy of 7B sizes. As larger models have less uncertainty and a clearer understanding of document passages, it helps to amp up internal representations of the documents.

\begin{table}[h]
    \centering
    \small
    \def\arraystretch{1.0}
    \caption{Accuracy of bigram medium size MLP for 100 Wikitext documents for \textit{test-in} and \textit{test-out} splits. The accuracy drop is natural as \textit{test-out} is unseen tokens. For all model sizes, the bigram cases show better generalization and unigram cases show the smallest accuracy drop.}
\begin{tabular}{c|ccc|ccc} 
\noalign{\hrule height 1pt}
 & \multicolumn{3}{|c}{Test-In} & \multicolumn{3}{|c}{Test-Out}  \\
      \noalign{\hrule height 1pt}
  LLM & unigram & bigram & trigram & unigram & bigram & trigram   \\  \hline
  \noalign{\hrule height 1pt}
llama2 13b & $0.919$ & $\mathbf{0.941}$ & $0.920$ & $0.791 (\textcolor{red}{ -0.129 })$ & $\mathbf{0.804}(\textcolor{red}{ -0.137 })$ & $0.592(\textcolor{red}{ -0.328 })$  \\
llama2 7b & $0.904$ & $\mathbf{0.931}$ & $0.902$ & $0.760 (\textcolor{red}{ -0.144 })$ & $\mathbf{0.777} (\textcolor{red}{ -0.154 })$ & $0.578 (\textcolor{red}{ -0.323 })$  \\
llama2-chat 13b & $0.908$ & $\mathbf{0.935}$ & $0.912$ & $0.784 (\textcolor{red}{ -0.124 })$ & $\mathbf{0.801} (\textcolor{red}{ -0.134 })$ & $0.587 (\textcolor{red}{ -0.325 })$  \\
llama2-chat 7b & $0.901$ & $\mathbf{0.928}$ & $0.901$ & $0.760 (\textcolor{red}{ -0.141 })$ & $\mathbf{0.777} (\textcolor{red}{ -0.150 })$ & $0.582 (\textcolor{red}{ -0.318 })$  \\
  \noalign{\hrule height 1pt}
    \end{tabular}
    
    \label{table:test_out}
\end{table}

\subsubsection{Logits as Inputs}
In the previous watermark work\cite{liu2023watermarking}, authors watermark machine-generated contents by modifying the next word probability. Similarly, we can use logits to tag documents. However, the number of dimensions for logits is larger than the hidden dimension size, resulting in a slow training time. Instead, we use the last hidden, a compact representation of the logits. Table \ref{table:logit_comparison} shows the test-in performance for logits with bigram medium-size MLP. We observe that the last hidden representation shows better generalization performance than logits.

\begin{table}[h]
    \centering
    \def\arraystretch{1.0}
    \caption{\textit{test-in} accuracy of bigram medium size MLP for 100 Wikitext documents. In all cases, the last hidden shows better generalization. }
    \begin{tabular}{c|c|c}
      \noalign{\hrule height 1pt}
  LLM & Hidden & Logit \\  \hline
  \noalign{\hrule height 1pt}
        Llama2 13b &  0.942  &  0.937   \\
        Llama2 7b &  0.935  &  0.915  \\
        Pythia 6.9b &  0.941  &  0.935   \\
        OPT 6.7b &  0.948  &  0.935  \\ 
  \noalign{\hrule height 1pt}
    \end{tabular}
    
    \label{table:logit_comparison}
\end{table}

\section{Discussion}

The source identifiers are post hoc modules for LLMs that do not interrupt the trained parameters and link representations to the source of documents. The only required resource is a single MLP for each token, which maps the representations of LLM to document labels. Such a post hoc module for LLM is essential to promote safe uses of LLMs. 

Text and document pairs are required to train the identification. This data pair is not accessible by LLM's end users. However, LLM providers can access the data. Therefore, the proposed method is mainly targeted at providers. We believe the process of giving source information to the end users can benefit both by increasing the reliability of the LLMs. For example, providing the origin of the text generated by ChatGPT can give the users additional information to double-check the factuality.

Our identification demonstration highlights the necessity of document scalability. Identification aims to provide a reliable source of the generated content. To do this, training on extreme labels is imperative. With the increasing number of documents, efficient training methods must be studied in the future. Note that the source identifier gives the document labels only in the list of labels; therefore, the obtained document index is limited. We believe additional work on extreme labels can trace the extreme documents used for LLMs. 

Another limitation of source identifiers is the possibility of false positives in documents, where irrelevant passages may be more likely to be identified. To mitigate this, more concrete and robust predictors are required. For example, instead of predicting one-hot labels, predicting the vector representation of a document can be a good inductive bias. 

We observe that the internal representations of LLMs are expressive enough to identify the text. The internal representations are progressively updated to make safe AI \cite{bai2022training}. As a result, the model can forget the document contents trained in the pertaining corpus. That is, the original representation of a document is transformed. Unlike this, humans can recall book names even in the continual learning setting. This study also suggests the need for human-level document content management for LLMs.

\section{Conclusion}

In this work, we introduce a token-level source identification problem that maps token representations in LLM to document labels. We verify that LLMs scalability holds for the generalization performance of the source identification. To increase the generalization, we propose bigram representations for the source identifier. Our extensive experiments on various LLM types, sizes, and locations consistently show that identification at the token level can provide the source of generation. Our work can inspire the safe use of LLMs, including copyright protection and reliability, by identifying the document sources.  

\section{Ethical Statements}
This work studies tracing the source of documents for generated text without interrupting the generation process. 
The LLM developers can provide the document sources to the end users with the proposed method. This process can increase the reliable usage of LLMs.  

\section{Acknowledgement}
This work was partly supported by Institute of Information $\&$ Communications Technology Planning $\&$ Evaluation (IITP) grant funded by the Korea government (MSIT) (No. 2022-0-00984, Development of Artificial Intelligence Technology for Personalized Plug-and-Play Explanation and Verification of Explanation; No. 2019-0-00075, Artificial Intelligence Graduate School Program (KAIST); No. 2022-0-00184, Development and Study of AI Technologies to Inexpensively Conform to Evolving Policy on Ethics), and Samsung Electronics MX Division.

%
%
%
%
%
\bibliographystyle{splncs04}
\bibliography{bibliography}
\end{document}